\documentclass{article}

\usepackage{PRIMEarxiv}

\usepackage[utf8]{inputenc} 
\usepackage[T1]{fontenc}    
\usepackage{hyperref}       
\usepackage{url}            
\usepackage{booktabs}       
\usepackage{amsfonts}       
\usepackage{nicefrac}       
\usepackage{microtype}      
\usepackage{lipsum}
\usepackage{fancyhdr}       
\usepackage{graphicx}       
\usepackage{graphicx}
\usepackage{float}

\usepackage{amssymb}
\usepackage{enumitem}
\usepackage{multirow}
\usepackage[table,xcdraw]{xcolor}

\usepackage{algorithm}
\usepackage{algpseudocode}
\usepackage{amsmath}
\usepackage{csquotes}
\usepackage{color}

\graphicspath{{media/}}     

\title{Video-based fully automatic assessment of open surgery suturing skills
}

\author{
  Adam Goldbraikh \\
  Applied Mathematics Department \\
  Technion – Israel Institute of Technology \\
  Haifa 3200003, Israel\\
  \texttt{sgoadam@compus.technion.ac.il} \\
   \And
  Anne-Lise D'Angelo \\
  Mayo Clinic, Surgery \\
  Rochester, MN, USA \\
  \texttt{annelise.dangelo@gmail.com} \\
   \AND
   Carla M. Pugh \\
   Stanford University, School of Medicine \\
   Stanford, CA, USA.\\
   \texttt{cpugh@stanford.edu} \\
   \And
   Shlomi Laufer \\
   Faculty of Industrial Engineering and Management\\
   Technion – Israel Institute of Technology \\
   Haifa 3200003, Israel \\
   \texttt{laufer@technion.ac.il} \\
}

\begin{document}
\maketitle

\begin{abstract}
\textbf{Purpose:} The goal of this study was to develop new reliable open surgery suturing simulation system for training medical students in situation where resources are limited or in the domestic setup. Namely, we developed an algorithm for tools and hands localization as well as identifying the interactions between them based on simple webcam video data, calculating motion metrics for assessment of surgical skill\\
\textbf{Methods:}
Twenty-five participants performed multiple suturing tasks using our simulator.The YOLO network has been modified to a multi-task network, for the purpose of tool localization and tool-hand interaction detection. This was accomplished by splitting the YOLO detection heads so that they supported both tasks with minimal addition to computer run-time. Furthermore, based on the outcome of the system, motion metrics were calculated. These metrics included traditional metrics such as time and path length as well as new metrics assessing the technique participants use for holding the tools.\\
\textbf{Results:}
The dual-task network performance was similar to that of two networks, while computational load was only slightly bigger than one network. In addition, the motion metrics showed significant differences between experts and novices.\\
\textbf{Conclusion:}
 While video capture is an  essential part of minimal invasive surgery, it is not an integral component of open surgery. Thus, new algorithms, focusing on the unique challenges open surgery videos present, are required.  In this study a dual-task network was developed to solve both a localization task and a hand-tool interaction task. The dual network may be  easily  expanded  to  a  multi-task  network,  which may be useful for images with multiple layers and for evaluating the interaction between these different layers.
\end{abstract}

\begin{LARGE}
	\begin{center}
		\color{gray}
		
		*Accepted at IJCARS*\\
	\end{center}

\end{LARGE}

\section{Introduction}
\label{sec:introduction}
Since Dr. Erich Mühe performed the first laparoscopic cholecystectomy in 1985, it has become the gold standard surgical treatment for gallbladder disease \cite{reynolds2001first}. Nevertheless, conversion to open surgery may be necessary in cases of medical and surgical complications related to anesthesia, peritoneal access, pneumoperitoneum, and thermocoagulation \cite{genc2011necessitates}. Cholecystectomy is just one of many examples of open surgery procedures that are being replaced by minimal invasive surgery (MIS), yet may require reverting to open surgery in the face of complications. Thus, while the new generation of surgeons has less experience with open surgery procedures \cite{eckert2010changing,mccoy2013open}, they still must master open surgery skills to handle the more extreme situations \cite{fonseca2013open}.

The recent advances in deep learning and computer vision have led to a growing number of studies focusing on automatic analysis of surgical video data \cite{twinanda2016endonet,al2018monitoring}. Since the use of video is an integral part of MIS, most of these studies have focused on laparoscopic and robotic surgery. In contrast, video capture is not well established in open surgery \cite{saun2019video}. Thus, open surgery has not benefited from the many advantages computer vision and deep learning methods have to offer for skill training and automatic assistance. Furthermore, while the use of simulators is inherent to training MIS \cite{gallagher2011fundamentals}, assessing open surgery skills, which is no less important, is lagging \cite{davies2013open,eckert2010changing}. This study will focus on both the development of novel video analysis algorithms as well as using these algorithms for assessment of surgical skills. 

There are some fundamental differences between the video data obtained during MIS and during open surgery. In MIS the video image usually includes 1-2 tool tips which are in actual use. Therefore, \emph{tool presence detection} and \emph{tool localization} are common goals in multiple studies analyzing MIS \cite{twinanda2016endonet,hu2017agnet}. Tool presence detection refers to identifying the existence of the tool in the image while tool localization indicates providing its position as well. In contrast, the video image during open surgery will often show not only the tool tip but also the hand. The image may include 2-4 hands captured concurrently in a range of positions and activities as well as stationary tools not held by anyone. For example, one surgeon might have a needle driver loaded with a needle in one hand and forceps lifting the tissue in the other hand. Meanwhile, another surgeon may be assisting by stabilizing the tissue with one hand and holding scissors in the other. Generating just a list of all the objects present in the image ignores the interaction between the objects and thus provides only a partial description of the image. Full analysis of the image structure should include the identification of the different tools and hands as well as their interactions.

The traditional teaching and assessment of technical skill relies heavily on the apprentice model, in which residents perform the procedure on a patient in the operation room (OR) under the guidance and evaluation of an expert. This approach does not provide a standardized method for training and assessing surgical skills \cite{darzi1999assessing}. This led to the development of simulation-based training and assessment. Traditionally, this included observer-generated task-specific checklists and global rating scales. Both methods are time-consuming and tend to bias \cite{d2015idle}. The crucial need for objective methods has motivated the development of technology-based approaches \cite{moorthy2003objective,reiley2011review,reznick2006teaching,d2015idle}. 

In recent years there is a growing interest in methods for tele-education and tele-simulation \cite{Perez2021Teleeducation,Roach2021Telesimulation}. Furthermore, with the recent outbreak of the Coronavirus (COVID-19) pandemic, the need for novel methods of remote education in general and the training of surgeons in particular has become clearer than ever \cite{luck2021undergraduate,siddiqui2020new,garcia2020image}. Sensor-enabled simulations may be integrated with remote education, thus providing objective assessment and feedback. Yet, they typically require expensive equipment and a complex setup, which is more appropriate for modern simulation centers and not for the home environment. This need to develop reliable surgical simulations that use cheap technology is not a new concern; it has been coming up in the context of developing countries where the resources are limited \cite{Lewis2019CognitiveAA,campain2018evaluation,hasan2019need,yadav2021telemedicine}.
Therefore, in this study we will evaluate a system that enables self-training and assessment of open surgery technical skills at the home of the trainee. With such a setup in mind, we captured video data using a standard webcam connected to a laptop. The algorithms developed are fast and can be analyzed on the cloud or even locally on the CPU or GPU within a reasonable processing time, providing evaluation scores in a timely manner. The simulator used includes a simulation board and basic surgical tools which can be supplied by mail.

The technical goal of this work is two-fold: first, study both surgical tool localization and surgical image structure; second, evaluate video-based kinematic analysis of technical surgical skills.
The contributions of our work are the following. We developed a variable tissue simulator which we use for assessment of open surgery skill \cite{d2015idle,d2016working}. Task analysis of the video revealed that open surgery video data require new categories which were not defined in previous studies on MIS video data. The task analysis was followed by the development of a new near-real time multi-task detection network for detecting the position of all the tools and hands in the image as well as identifying which tool is being used by each hand. We used the output of these algorithms to assess surgical performance based on multiple motion metrics. Finally, our multi-task system provides hand location, tool location, and hand-tool interactions. This combined knowledge led to the development of a new motion metric that examines the technique used for holding the tool.
\begin{figure}
	\centering{
		\includegraphics[scale=0.3]{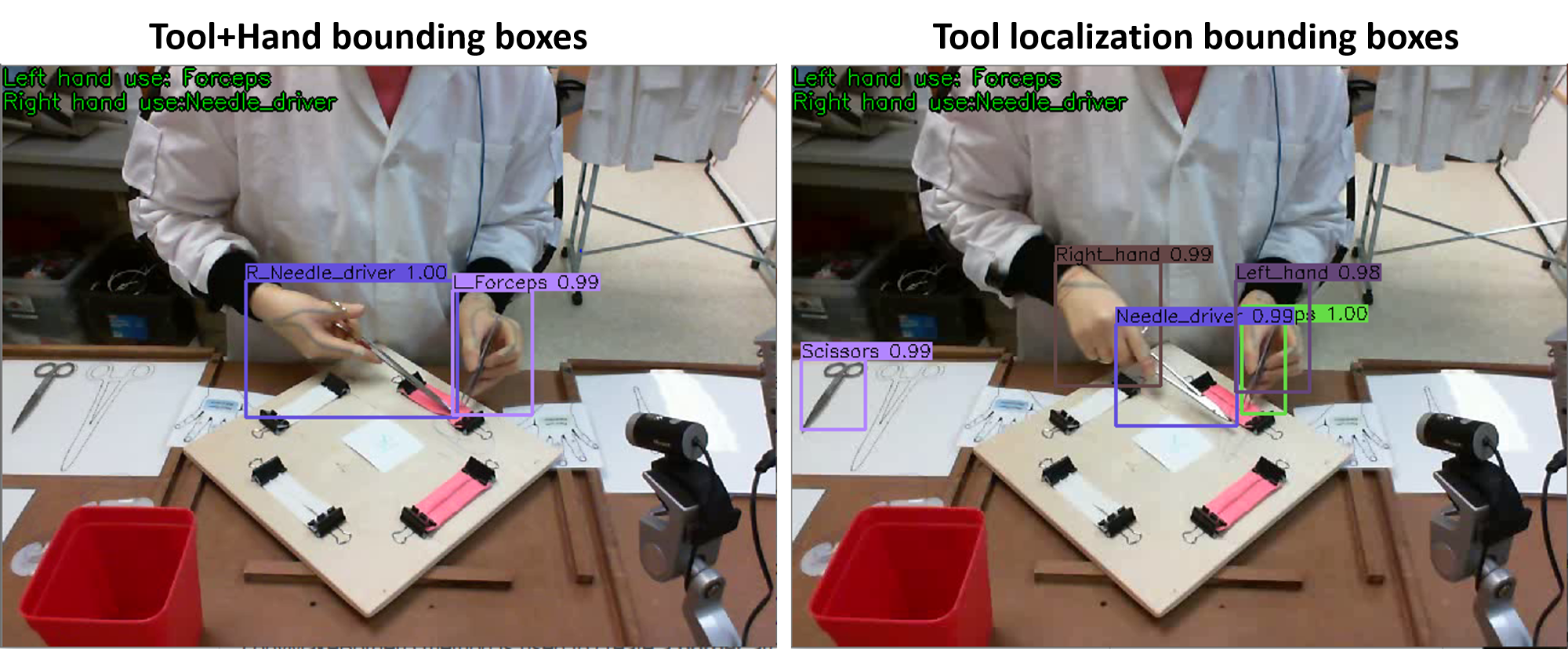}
	}
	\caption{\label{Fig:Fig1} Each bounding box in the left figure represents a hand-tool interaction object, and every box in the right figure represents tool or hand. The text in the top-left corner outputs the final decision on the tool usage.}
\end{figure}

\section{Related Works}

Multiple studies have demonstrated that kinematic data can provide valuable information in the assessment of surgical skill \cite{moorthy2003objective,reiley2011review,reznick2006teaching}. However, in open surgery, most studies use kinematic data from sensors such as Electromagnetic 6DOF Sensors for the measurement of hand motion \cite{d2016working,moorthy2003objective,reiley2011review,reznick2006teaching}. These sensors are typically expensive and may have a complex setup as well as interfere with the normal workflow. Based on the kinematic data, different skill-evaluation metrics are calculated, such as procedure time, path length, number of hand movements, working volume, etc. \cite{d2015idle,d2016working}. Each metric can indicate different aspects of motor skills level; in this sense these types of models are highly explainable.

Simulators measuring kinematic data have been developed for robotic and minimally invasive surgery as well \cite{reiley2011review}. Several public data-sets contain robotic kinematic data with skills assessment labeling such as JIGSAWS and MISTIC-SL \cite{gao2014jhu,gao2014language}. Machine learning methods that predict the surgeon's level of expertise based on these metrics have been developed \cite{fard2016machine,zia2018automated}, as well as deep learning methods that predict the level of expertise directly on the data (kinematic or video) without intermediate feature calculation \cite{fawaz2018evaluating,funke2019video}. Kinematic data can be obtained also by using computer vision methods upon video data, e.g. by using bounding boxes of object detection \cite{jin2018tool} or landmark detection \cite{du2018articulated} on hands or tool tips. 

Analysis of MIS video data has raised a range of research questions. Evaluation of topics such as tool presence detection \cite{katic2014knowledge,jin2018tool}, workflow recognition \cite{cleary2004or2020,ramesh2021multi}, error identification and skill assessment \cite{partridge2014accessible,jin2018tool} has the potential of making the surgical environment safer and more efficient. In a recent study, detection of hands in open surgery videos was assessed \cite{zhang2020using}.

	The goal of this study is to detect tools and hands. Therefore, the selection of the optimal object detection algorithm, which is the engine of our system, is of the utmost importance. Current object detection algorithms are generally grouped into two main families: two-stage algorithms and one-stage algorithms. In the two-stage algorithms, the first stage extracts regions of interest, namely those regions where the objects are expected to be found. The second stage involves classifying the objects and locating their bounding boxes. Two stage algorithms, such as Faster R-CNN \cite{ren2015faster} and Mask R-CNN \cite{he2017mask}, are characterized by high accuracy rates and long run time. As a result, they are not appropriate for real-time applications.
	In contrast, the one-stage object detection algorithms do not require intermediate steps, as they frame the object detection as a single regression for both identifying bounding boxes and classifying them in one stage. These algorithms are less accurate than the two-stage algorithms, but because they work much faster than the two-stage algorithms, they are more suitable for real-time applications. This family includes the SSD algorithm \cite{liu2016ssd}, RetinaNet \cite{lin2017focal} and all versions of YOLO \cite{redmon2016you,redmon2017yolo9000,redmon2018yolov3}.
	A new sub-family of one stage object detection algorithms has been recently introduced. These algorithms are based on the Transformer architecture and involve a single stage. While this renders them more accurate, they are slower and therefore not suitable for real-time applications \cite{carion2020end,liu2021Swin}.
	In general, if fast and accurate detection is required, the one-stage YOLOv3 is considered as a great choice. For example, YOLOv3 was used for real-time Jellyfish classification \cite{gao2021real}, real-time people detection \cite{hassan2020people}, and real-time pattern-recognition of ground-penetrating radar images \cite{li2020real}.
	In the surgical tool detection domain, two-stage object detection algorithms have been implemented, such as R-CNN based networks \cite{jin2018tool} as well as one-stage algorithms where inference time is significant, such as YOLO9000 in \cite{jo2019robust} and RetinaNet in \cite{zhang2020using} and SSD in \cite{ali2019object}.

There are some topics in the computer-vision community that can be considered as related to the unique challenges of open surgery, such as Hand-Object Interaction \cite{schroder2017hand}, Human-Object Interaction \cite{li2019transferable} and Object-Object Interaction\cite{herzig2019spatio}.
In \cite{herzig2019spatio} the authors propose inter-object graph representation for recognition of activities in self-driving scenarios. Their method is based on disentangled graph embedding with direct edge appearance observation. Their most relevant observation to our work is that relations between objects are captured in a single bounding box that contains both interacted objects strongly rather than in using tight boxes of the objects separately.

\section{Methods}

The first goal of this work was to detect the position of all the tools and hands in the image; that is, provide \emph{tool localization}. The training set contained images with labeled objects and their locations. We specified the object's location by bounding it with a tight box. In the labeled images, all the tools and all the hands (whether holding a tool or not) were outlined. We used a YOLO detection network \cite{redmon2018yolov3} for the \emph{tool localization} task. It should be noted that in some cases the hand is empty, for example when palpating tissue. In this case the hand may be regarded as the “tool.” However, in most cases the hand is holding a tool. For simplicity, when we use the term \emph{tool localization}, we mean all the tools and all the hands in the image.

The second goal was to determine the \emph{hand-tool interaction}. In a previous study \cite{Goldbraikh2020} we used the output of the \emph{tool localization} algorithm combined with spatial assumptions to match between the hand and the tool. For each hand detected we examined whether there was a tool in close proximity and if so, it was assumed it was being used by that hand. The labeling for such a task was simpler. We only needed to annotate start and end time of each tool usage. Therefore, for this task we labeled the full data-set (whereas for \emph{tool localization}, we only labeled a subset of the images, as in other deep learning studies). 

However, this approach is based on heuristic assumptions and prone to manual fine tuning. In this study we will develop a standard approach using deep learning to provide a general solution. For this we added another layer of annotation. In each image annotated for the \emph{tool localization} task we added another set of bounding boxes. In this new set, each pair of hand+tool is outlined using a tight box (Fig. \ref{Fig:Fig1}). Now we can use a detection network such as YOLO for determining the \emph{hand-tool interaction} as well. It should be noted that now we have two sets of ground truth labels for the \emph{hand-tool interaction} task. The first includes the entire video set; however, it includes only the interaction and not its spatial position. The second set includes only a small sample of the images. This set includes bounding boxes and is used for training and testing using traditional machine learning approaches.
There are two naïve methods to train the network to provide the complete image structure (\emph{tool localization} and \emph{hand-tool interaction}). The first is to train one network for detecting the tools and another, separate network for detecting the combination of \emph{hand-tool interaction}. As we will show in the results, this method provides a good outcome; however, analysis requires twice the computation power. The second approach would be to combine all the labeled data (the tools, the hands and the hand+tool) then train one network with the entire data set. This approach saves computation power; however, as we will see, it leads to a significant decrease in accuracy.

Therefore, in this study we developed a multi-task detection network to solve both (\emph{tool localization} and \emph{hand-tool interaction}) challenges. The multi-task network is based on the YOLO network; however, we updated the final layers to support multiple detection tasks. Using this new network, we gain from both worlds: our detection is as good as two separate networks while the required computation power is only slightly more than one network. 

\subsection{Variable tissue simulator}

The variable tissue simulator was developed to simulate a suturing task to assess decision making during suturing tasks of varying difficulty. The simulator consists of a board to which simulated material is connected by two clips \cite{d2015idle}. The task was to place three interrupted instrument-tied sutures on two opposing pieces of the material. Two different materials were used: tissue paper simulating friable tissue, and rubber balloons simulating arteries. Each participant was provided with three tools: a needle driver, surgical forceps, and suture scissors. Top view video data were captured in a frame rate of 30 FPS.



\subsection{Data Collection}\label{section:Data_Collection}
Eleven medical students, one resident, and 13 attending surgeons participated in the study. Each participant performed twice on the friable tissue simulator and twice on the artery simulator. Thus, there were a total of 100 videos, each approximately 2-6 minutes long. This data-set was split into two sets. Where the first contains 15 videos, we will refer to it as \emph{train video set} and the second, \emph{test video set}, contains the rest.
From the train video set $924$ frames were picked and split to seven sub-sets of 132 images for k-fold Cross-Validation. Each selected frame was labeled with two sets of bounding boxes. The first for the \emph{tool localization} task and the second for the \emph{hand-tool interaction} task.
In addition, from 5 other videos, 200 frames were chosen for a \emph{test set}. These frames were labeled in the same method. The labeling was performed with Microsoft’s Visual Object Tagging Tool. Finally, the entire video data-set was labeled for the start and end time of each tool usage using Behavioral Observation Research Interactive Software (BORIS).
As mentioned, we have two sets of ground truth labeled data. The first includes tight bounding boxes and is used for the training and testing of the different classifiers. In this set we require an $IoU$ of at least 0.5 with the ground truth bounding box to be considered as a true prediction, and average precision (AP) is used to assess the results. The second labeled data set includes all the data; however, it does not include any bounding boxes, only start and end point. Therefore, this set is only used for testing the \emph{hand-tool interaction} performance. This is assessed using Precision, Recall, and F1 metrics.

The following data augmentation was used during the training. Horizontal flip applied with a probability of $0.5$. 
We uniformly randomly rotated the images and automatically fit corresponding bounding boxes, based on its transformation matrix in the range of $\pm 7 ^{\circ}$. In addition, we use the standard Pytorch Torchvision ColorJitter module, which randomly changes the color values with the following parameters: brightness=0.2, contrast=0.2, saturation=0.1, and hue=0.05.
The training and evaluation were performed on a NVIDIA Tesla V100 Volta GPU Accelerator 32GB Graphics Card.

\subsection{Naive Approaches}

Two naive approaches may be used for solving the \emph{tool localization} and \emph{hand-tool interaction} problems. The first approach is to train two separate networks, one for each problem. One network will include five categories of tool types - $D$ and the second the eight classes of \emph{hand-tool interaction} combinations - $S$. The second approach is to train one integrated system for both problems. In this case the merged fourteen categories - $D \cup S$ will be used. For all tasks a YOLOv3 (YOLO) network was used. 

All three systems were trained under similar conditions: Adam optimizer up to 400 epochs with learning rate of $10^{-3}$ and additional 200 epochs with learning rate of $10^{-4}$. The model selected was the model with the highest AP with respect to the validation set. During the test session, each frame was tested twice, the first time with a horizontal flip and the second time without.
The models are tested on the $200$ test frames as defined in the previous section.

\subsection{Multi-Task Deep Neural Network Approach}
\begin{figure}
	\centering{
		\includegraphics[scale=0.19]{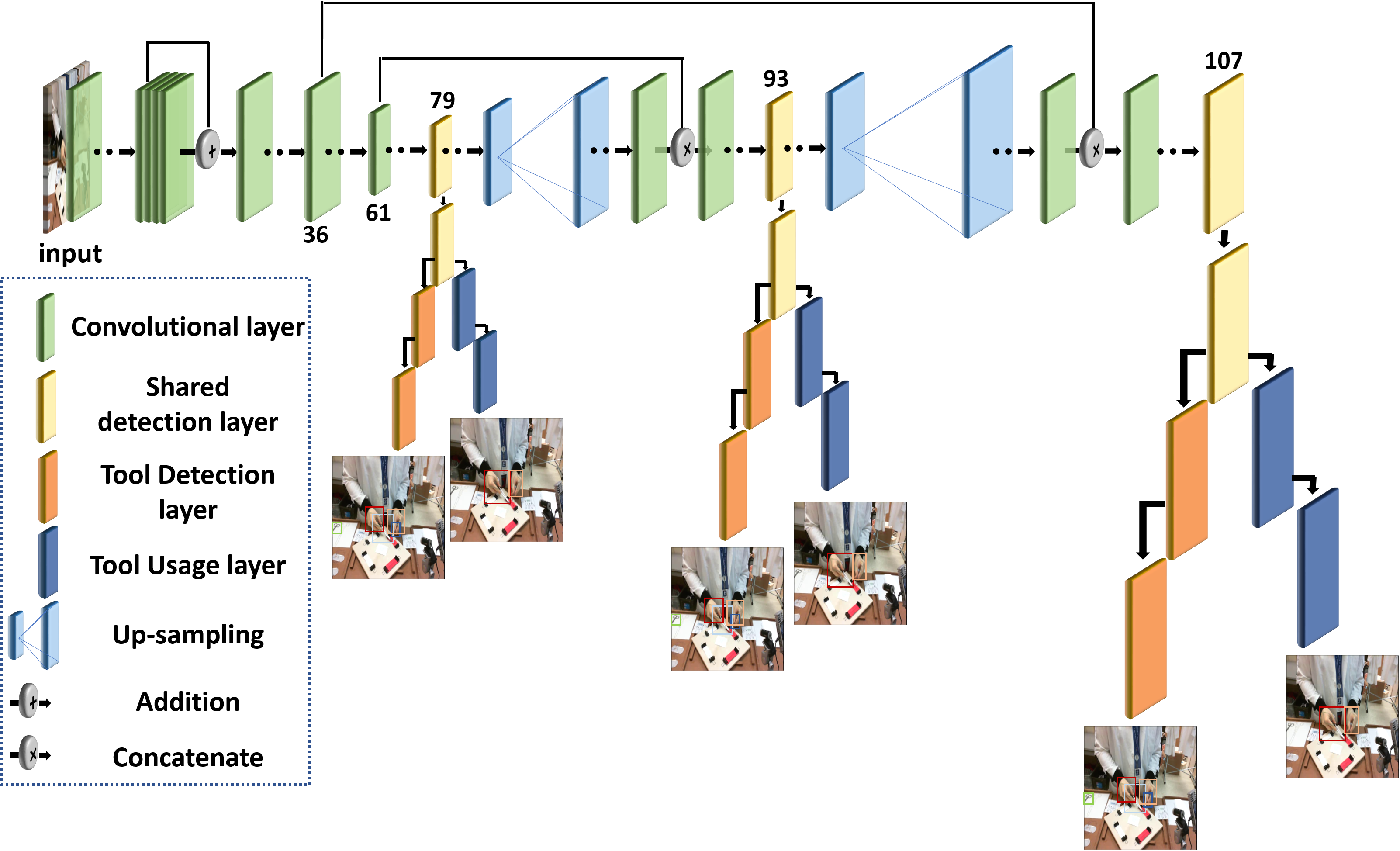}
	}
	\caption{\label{Fig:net} Schematic illustration of proposed architecture}
\end{figure}

\begin{figure}
	\centering{
		\includegraphics[scale=0.19]{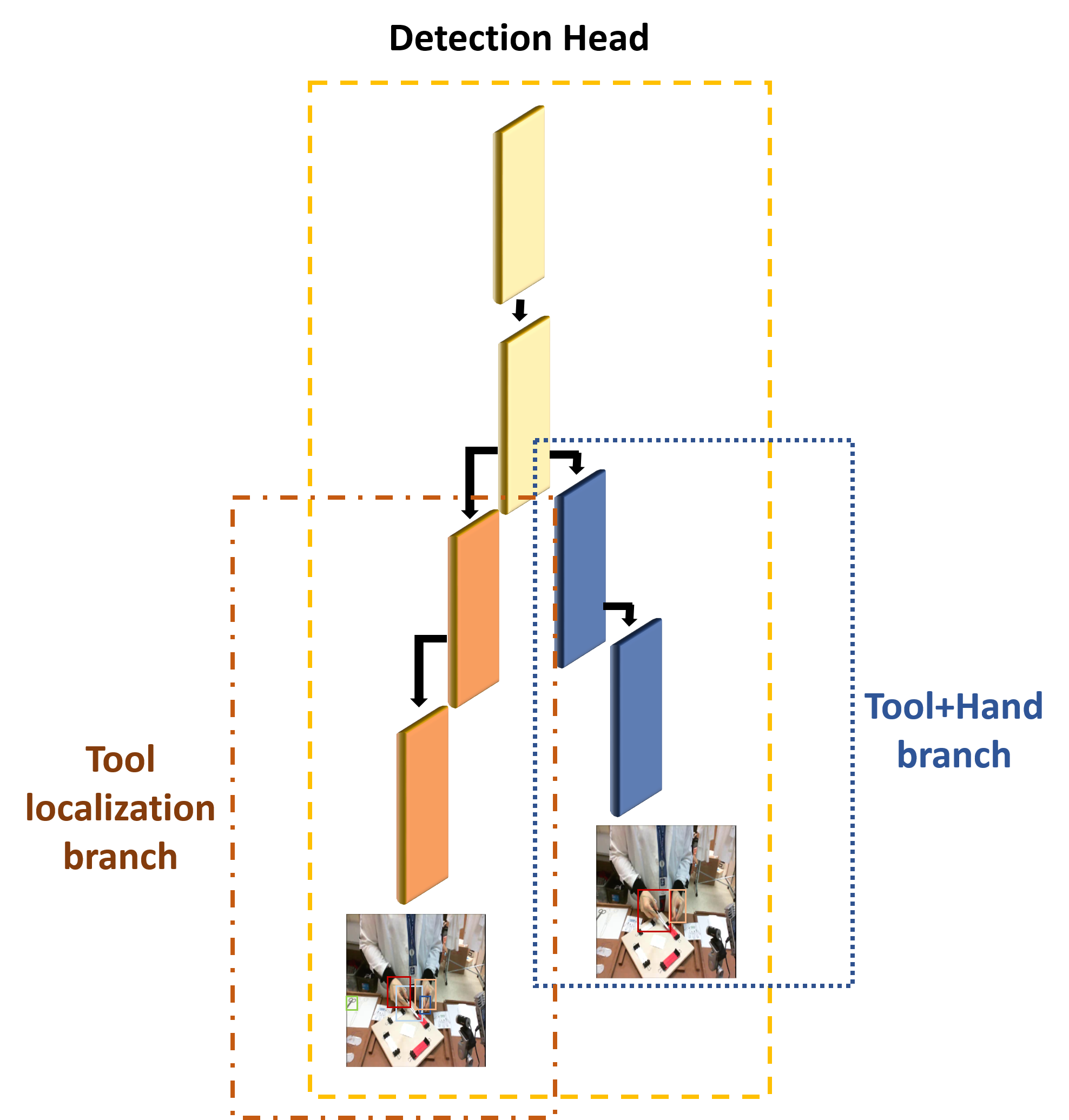}
	}
	\caption{\label{Fig:DetectionHead} The split in the detection head to \emph{tool localization} branch and \emph{hand-tool interaction} branch}

\end{figure}

Our architecture is an extension of the YOLO network \cite{redmon2018yolov3}. YOLO is a fully convolutional network that consists of 106 layers. To provide detection at multiple image scales YOLO uses multiple detection heads. Each head consists of four convolutional layers and one YOLO prediction layer. The input image size is $416 \times 416$, and three color channels are used. In our new architecture each detection head was split after the second convolutional layer. We will refer to the layers after the split as branches. Hence, every original detection head was split into two branches: a \emph{hand-tool interaction} branch and a \emph{tool localization} branch (see Fig. \ref{Fig:DetectionHead}). The output of each branch is shown in Figure \ref{Fig:Fig1}. We will refer to all layers that are accessible to both branches as the trunk of the network. The new architecture is depicted in Fig. \ref{Fig:net}. YOLO uses \emph{Non-maximum Suppression} (NMS) to address the problem of multiple detections of the same object. For the \emph{tool localization} branch an additional NMS was added. It suppressed multiple tools in the same area while allowing for tools and hands to overlap. 

\paragraph{Training Method:}
Technically, we have two data-sets, one for the \emph{tool localization} task and the other for the \emph{hand-tool interaction} task. Note, that the two data-sets are based on the same set of images and differ only by the tagging.
Each epoch contains batches from the two data-sets. The batches are ordered in round-robin fashion, where after each \emph{tool localization} batch was placed, a \emph{hand-tool interaction} batch was used. While training for one task, the branch dedicated to the other task was frozen. The network was trained by using Adam optimizer with a learning rate of $10^{-3}$ for the first 200 epochs and a learning rate of $10^{-4}$ for an additional 100 iterations.We trained in k-fold Cross-Validation manner (K=7), where every model, in addition to its validation, also tested on our test-set.

\paragraph{Inference of Tool Usage:}
The goal of the \emph{hand-tool interaction} was to provide exactly one bounding box for each hand visible in the image. When unsuccessful, data from the \emph{tool localization branch} may be used to help infer tool usage. This includes two scenarios (see Fig. \ref{Fig:Scenarios}):

\begin{figure}
	\centering{
		\includegraphics[scale=0.5]{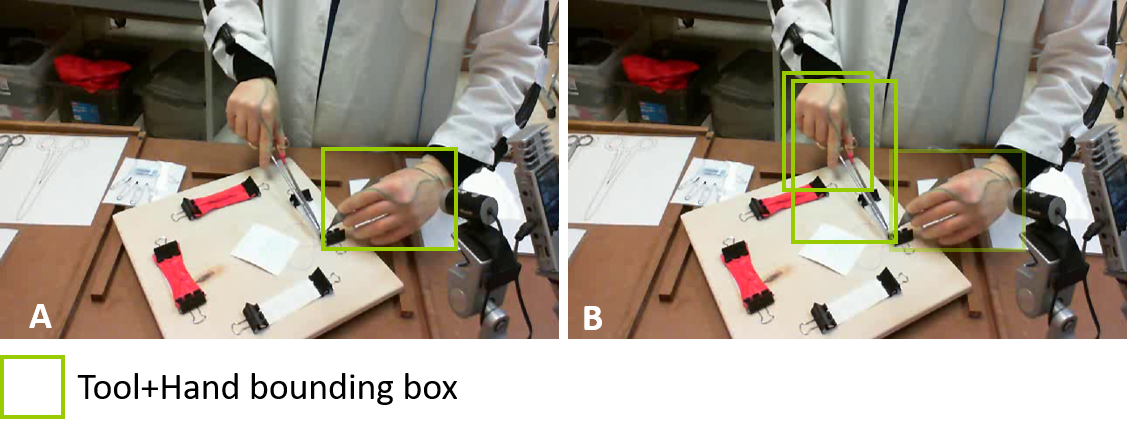}
	}
	\caption{\label{Fig:Scenarios} A - Scenario 1: No \emph{hand-tool interaction} bounding box for the right hand, B - Scenario 2: There are two \emph{hand-tool interaction} bounding box for the right hand}

\end{figure}

\paragraph{Scenario 1- No bounding box:}
This scenario includes situations in which the \emph{hand-tool interaction} branch provides no bounding box for one of the hands, yet the \emph{tool localization} branch detects that hand. In this case, we search the output of the \emph{tool localization} branch for an overlap between the hand's bounding box and one of the tool's bounding boxes, described more specifically in \cite{Goldbraikh2020}.

\paragraph{Scenario 2- Multiple bounding boxes:}
This scenario includes situations in which the \emph{hand-tool interaction} branch provides multiple bounding boxes of one of the hands, and we need to select the correct bounding box. All the bounding boxes for that hand will be compared with all the bounding boxes of all the tools detected by the \emph{tool localization} branch; the pair with the largest overlap will be selected.

\paragraph{Smoothing process:}
For each frame the final \emph{tool-hand interaction} was based on the majority of the previous 15 frames. In addition, empirical data revealed that when the hands were moving fast, \emph{tool localization} was significantly reduced due to image blurriness while hand detection was not reduced. Therefore, hand speed was calculated based on hand detection data. Decisions regarding \emph{tool-hand interaction} were not changed during fast movement. The smoothing process only influenced the decision of which tools were being used at any given moment and did not affect the bounding boxes provided by the system.

\section{Results}
\subsection{Results Naive Approaches}
Table \ref{table:table1} summarize the results for the naive approaches. The tool localization only network yielded mAP of $0.863$ and the integrated network achieved only $0.758$ on these classes. The Tool+Hand only network has mAP of $0.871$ on it's 8 classes when the combined network has only $0.621$. In total, the separate networks show significantly better results, with mAP of $0.868$ than the one integrated system, with mAP of $0.674$.

\begin{table}[h]
	{\centering \scalebox{.59}{
			
			
			\begin{tabular}{l|c|l|l}
				\hline
				\rowcolor[HTML]{DAE8FC} 
				\multicolumn{1}{|l|}{\cellcolor[HTML]{DAE8FC}}                                                                                 & \textbf{\begin{tabular}[c]{@{}c@{}}Tool localization\\  ($AP_{50}$)\end{tabular}}          & \multicolumn{1}{c|}{\cellcolor[HTML]{DAE8FC}\textbf{\begin{tabular}[c]{@{}c@{}}Tool+Hand\\  ($AP_{50}$)\end{tabular}}} & \multicolumn{1}{c|}{\cellcolor[HTML]{DAE8FC}\textbf{\begin{tabular}[c]{@{}c@{}}Regular Yolo (13)\\ with all classes\\  ($AP_{50}$)\end{tabular}}} \\ \hline
				\multicolumn{1}{|l|}{\textbf{Right hand}}                                                                                      & \multicolumn{1}{l|}{0.956}                                                         & \multicolumn{1}{c|}{}                                                                                              & \multicolumn{1}{l|}{0.766}                                                                                                                     \\ \hline
				\multicolumn{1}{|l|}{\textbf{Left hand}}                                                                                       & \multicolumn{1}{l|}{0.947}                                                         & \multicolumn{1}{c|}{}                                                                                              & \multicolumn{1}{l|}{0.783}                                                                                                                     \\ \hline
				\multicolumn{1}{|l|}{\textbf{Needle driver}}                                                                                   & \multicolumn{1}{l|}{0.839}                                                         & \multicolumn{1}{c|}{}                                                                                              & \multicolumn{1}{l|}{0.740}                                                                                                                     \\ \hline
				\multicolumn{1}{|l|}{\textbf{Forceps}}                                                                                         & \multicolumn{1}{l|}{0.728}                                                         & \multicolumn{1}{c|}{}                                                                                              & \multicolumn{1}{l|}{0.659}                                                                                                                     \\ \hline
				\multicolumn{1}{|l|}{\textbf{Scissors}}                                                                                        & \multicolumn{1}{l|}{0.845}                                                         & \multicolumn{1}{c|}{}                                                                                              & \multicolumn{1}{l|}{0.843}                                                                                                                     \\ \hline
				\rowcolor[HTML]{C0C0C0} 
				\multicolumn{1}{|l|}{\cellcolor[HTML]{C0C0C0}\textbf{\begin{tabular}[c]{@{}l@{}}mAP\\ (Tool localization only)\end{tabular}}} & \multicolumn{1}{l|}{\cellcolor[HTML]{C0C0C0}{\color[HTML]{000000} \textbf{0.863}}} & \multicolumn{1}{c|}{\cellcolor[HTML]{C0C0C0}\textbf{}}                                                             & \multicolumn{1}{l|}{\cellcolor[HTML]{C0C0C0}{\color[HTML]{000000} \textbf{0.758}}}                                                             \\ \hline
				\multicolumn{1}{|l|}{\textbf{Scissors in right hand}}                                                                          &                                                                                    & 0.807                                                                                                              & \multicolumn{1}{l|}{0.731}                                                                                                                     \\ \hline
				\multicolumn{1}{|l|}{\textbf{Scissors in left hand}}                                                                           &                                                                                    & 0.827                                                                                                              & \multicolumn{1}{l|}{0.751}                                                                                                                     \\ \hline
				\multicolumn{1}{|l|}{\textbf{Needle driver in right hand}}                                                                     &                                                                                    & 0.927                                                                                                              & \multicolumn{1}{l|}{0.916}                                                                                                                     \\ \hline
				\multicolumn{1}{|l|}{\textbf{Needle driver in left hand}}                                                                      &                                                                                    & 0.907                                                                                                              & \multicolumn{1}{l|}{0.876}                                                                                                                     \\ \hline
				\multicolumn{1}{|l|}{\textbf{Forceps in right hand}}                                                                           &                                                                                    & 0.918                                                                                                              & \multicolumn{1}{l|}{0.854}                                                                                                                     \\ \hline
				\multicolumn{1}{|l|}{\textbf{Forceps in left hand}}                                                                            &                                                                                    & 0.954                                                                                                              & \multicolumn{1}{l|}{0.839}                                                                                                                     \\ \hline
				\multicolumn{1}{|l|}{\textbf{Empty right hand}}                                                                                &                                                                                    & 0.787                                                                                                              & \multicolumn{1}{l|}{0.001}                                                                                                                     \\ \hline
				\multicolumn{1}{|l|}{\textbf{Empty Left hand}}                                                                                 &                                                                                    & 0.838                                                                                                              & \multicolumn{1}{l|}{0.000}                                                                                                                     \\ \hline
				\rowcolor[HTML]{C0C0C0} 
				\multicolumn{1}{|l|}{\cellcolor[HTML]{C0C0C0}\textbf{mAP (Tool + Hand only)}}                                                   & \textbf{}                                                                          & \cellcolor[HTML]{C0C0C0}{\color[HTML]{000000} \textbf{0.871}}                                                      & \multicolumn{1}{l|}{\cellcolor[HTML]{C0C0C0}{\color[HTML]{000000} \textbf{0.621}}}                                                             \\ \hline
				\rowcolor[HTML]{9B9B9B} 
				\textbf{mAP (Both tasks)}                                                                                                      & \multicolumn{2}{c|}{\cellcolor[HTML]{9B9B9B}\textbf{0.868}}                                                                                                                                             & \textbf{0.674}                                                                                                                                 \\ \hline
		\end{tabular}}
		
	}
	\caption{\label{table:table1}%
		Average precision ($AP_{50}$) of each class in the tool localization network, Tool+Hand network, combined results of tool localization network and  Tool+Hand network, and integrated network with all 13 classes. The results are with respect to the  test set.
	}
	
\end{table}

\subsection{Results Multi-Task Deep Neural Network Approach}
The performance of the multi-task system in the \emph{tool localization} framework is depicted in Table \ref{table:table2}. The detection was slightly better than the two separate networks; this is a known effect of multitask training. The system yielded an expectation $mAP = 0.874$ for the \emph{tool localization} task, $mAP=0.885$ for the \emph{hand-tool interaction} task, and the overall results were $mAP = 0.881$. The results are based on the 200 \emph{test set} images. In the results table, one can see also the validation set results.

\emph{Tool-hand interaction} was also tested using the $85$ full videos from which frames were not taken for the training process. For $3.92\%$ of the data the \emph{hand-tool interaction} branch was unsuccessful and tool usage was inferred using the data from the \emph{tool localization} branch. The precision, recall, and $F_{1}$ of each tool as well their average values and the total accuracy were calculated (Table \ref{table:table3}).

\begin{table*}
	\centering
	\scalebox{.59}{
		
		\begin{tabular}{|c|c|c|c|c|c|c|c|c|c|}
			\hline
			& \multicolumn{9}{c|}{\textbf{Tool+Hand Detection}}                                                                                                                                                                                                                                                                                                                                                                                                                                                                                                                                                                                                                                                                                                                                                                                                    \\ \hline
			\multirow{2}{*}{\textbf{}}                                                               & \multirow{2}{*}{\textbf{\begin{tabular}[c]{@{}c@{}}Scissors in\\  right hand\end{tabular}}} & \multirow{2}{*}{\textbf{\begin{tabular}[c]{@{}c@{}}Scissors in\\  left hand\end{tabular}}} & \multirow{2}{*}{\textbf{\begin{tabular}[c]{@{}c@{}}Needle driver\\  in right hand\end{tabular}}} & \multirow{2}{*}{\textbf{\begin{tabular}[c]{@{}c@{}}Needle driver\\ in left hand\end{tabular}}} & \multirow{2}{*}{\textbf{\begin{tabular}[c]{@{}c@{}}Forceps in\\ right hand\end{tabular}}} & \multirow{2}{*}{\textbf{\begin{tabular}[c]{@{}c@{}}Forceps in\\  left hand\end{tabular}}} & \multirow{2}{*}{\textbf{\begin{tabular}[c]{@{}c@{}}Empty\\ right hand\end{tabular}}} & \multirow{2}{*}{\textbf{\begin{tabular}[c]{@{}c@{}}Empty\\ left hand\end{tabular}}} & \multirow{2}{*}{\textbf{mAP}}                                                    \\
			&                                                                                             &                                                                                            &                                                                                                  &                                                                                                &                                                                                           &                                                                                           &                                                                                      &                                                                                     &                                                                                  \\ \hline
			\multirow{2}{*}{\textbf{\begin{tabular}[c]{@{}c@{}}Validation\\  (AVG+SD)\end{tabular}}} & \multirow{2}{*}{\begin{tabular}[c]{@{}c@{}}0.896±\\ 0.035\end{tabular}}                     & \multirow{2}{*}{\begin{tabular}[c]{@{}c@{}}0.890±\\ 0.053\end{tabular}}                    & \multirow{2}{*}{\begin{tabular}[c]{@{}c@{}}0.955±\\ 0.028\end{tabular}}                          & \multirow{2}{*}{\begin{tabular}[c]{@{}c@{}}0.954±\\ 0.02\end{tabular}}                         & \multirow{2}{*}{\begin{tabular}[c]{@{}c@{}}0.898±\\ 0.04\end{tabular}}                    & \multirow{2}{*}{\begin{tabular}[c]{@{}c@{}}0.901±\\ 0.041\end{tabular}}                   & \multirow{2}{*}{\begin{tabular}[c]{@{}c@{}}0.909±\\ 0.020\end{tabular}}              & \multirow{2}{*}{\begin{tabular}[c]{@{}c@{}}0.91±\\ 0.02\end{tabular}}               & \multirow{2}{*}{\textbf{\begin{tabular}[c]{@{}c@{}}0.914±\\ 0.017\end{tabular}}} \\
			&                                                                                             &                                                                                            &                                                                                                  &                                                                                                &                                                                                           &                                                                                           &                                                                                      &                                                                                     &                                                                                  \\ \hline
			\multirow{2}{*}{\textbf{\begin{tabular}[c]{@{}c@{}}Test\\ (AVG+SD)\end{tabular}}}        & \multirow{2}{*}{\begin{tabular}[c]{@{}c@{}}0.849±\\ 0.045\end{tabular}}                     & \multirow{2}{*}{\begin{tabular}[c]{@{}c@{}}0.820±\\ 0.051\end{tabular}}                    & \multirow{2}{*}{\begin{tabular}[c]{@{}c@{}}0.926±\\ 0.009\end{tabular}}                          & \multirow{2}{*}{\begin{tabular}[c]{@{}c@{}}0.917±\\ 0.013\end{tabular}}                        & \multirow{2}{*}{\begin{tabular}[c]{@{}c@{}}0.915±\\ 0.018\end{tabular}}                   & \multirow{2}{*}{\begin{tabular}[c]{@{}c@{}}0.92±\\ 0.031\end{tabular}}                    & \multirow{2}{*}{\begin{tabular}[c]{@{}c@{}}0.87±\\ 0.011\end{tabular}}               & \multirow{2}{*}{\begin{tabular}[c]{@{}c@{}}0.864±\\ 0.014\end{tabular}}             & \multirow{2}{*}{\textbf{\begin{tabular}[c]{@{}c@{}}0.885±\\ 0.012\end{tabular}}} \\
			&                                                                                             &                                                                                            &                                                                                                  &                                                                                                &                                                                                           &                                                                                           &                                                                                      &                                                                                     &                                                                                  \\ \hline
			\textbf{}                                                                                & \multicolumn{9}{c|}{\textbf{Tool localization}}                                                                                                                                                                                                                                                                                                                                                                                                                                                                                                                                                                                                                                                                                                                                                                                                      \\ \hline
			\multirow{2}{*}{\textbf{}}                                                               & \multirow{2}{*}{\textbf{Right hand}}                                                        & \multirow{2}{*}{\textbf{Left hand}}                                                        & \multirow{2}{*}{\textbf{Needle driver}}                                                          & \multirow{2}{*}{\textbf{Forceps}}                                                              & \multirow{2}{*}{\textbf{Scissors}}                                                        & \multirow{2}{*}{\textbf{}}                                                                & \multirow{2}{*}{\textbf{}}                                                           & \multirow{2}{*}{\textbf{}}                                                          & \multirow{2}{*}{\textbf{mAP}}                                                    \\
			&                                                                                             &                                                                                            &                                                                                                  &                                                                                                &                                                                                           &                                                                                           &                                                                                      &                                                                                     &                                                                                  \\ \hline
			\multirow{2}{*}{\textbf{\begin{tabular}[c]{@{}c@{}}Validation\\  (AVG+SD)\end{tabular}}} & \multirow{2}{*}{\begin{tabular}[c]{@{}c@{}}0.932±\\ 0.012\end{tabular}}                     & \multirow{2}{*}{\begin{tabular}[c]{@{}c@{}}0.920±\\ 0.013\end{tabular}}                    & \multirow{2}{*}{\begin{tabular}[c]{@{}c@{}}0.937±\\ 0.016\end{tabular}}                          & \multirow{2}{*}{\begin{tabular}[c]{@{}c@{}}0.866±\\ 0.038\end{tabular}}                        & \multirow{2}{*}{\begin{tabular}[c]{@{}c@{}}0.959±\\ 0.028\end{tabular}}                   & \multirow{2}{*}{\textbf{}}                                                                & \multirow{2}{*}{\textbf{}}                                                           & \multirow{2}{*}{\textbf{}}                                                          & \multirow{2}{*}{\textbf{\begin{tabular}[c]{@{}c@{}}0.923±\\ 0.015\end{tabular}}} \\
			&                                                                                             &                                                                                            &                                                                                                  &                                                                                                &                                                                                           &                                                                                           &                                                                                      &                                                                                     &                                                                                  \\ \hline
			\multirow{2}{*}{\textbf{\begin{tabular}[c]{@{}c@{}}Test\\ (AVG+SD)\end{tabular}}}        & \multirow{2}{*}{\begin{tabular}[c]{@{}c@{}}0.937±\\ 0.01\end{tabular}}                      & \multirow{2}{*}{\begin{tabular}[c]{@{}c@{}}0.933±\\ 0.009\end{tabular}}                    & \multirow{2}{*}{\begin{tabular}[c]{@{}c@{}}0.856±\\ 0.014\end{tabular}}                          & \multirow{2}{*}{\begin{tabular}[c]{@{}c@{}}0.756±\\ 0.008\end{tabular}}                        & \multirow{2}{*}{\begin{tabular}[c]{@{}c@{}}0.890±\\ 0.016\end{tabular}}                   & \multirow{2}{*}{\textbf{}}                                                                & \multirow{2}{*}{\textbf{}}                                                           & \multirow{2}{*}{\textbf{}}                                                          & \multirow{2}{*}{\textbf{\begin{tabular}[c]{@{}c@{}}0.874±\\ 0.004\end{tabular}}} \\
			&                                                                                             &                                                                                            &                                                                                                  &                                                                                                &                                                                                           &                                                                                           &                                                                                      &                                                                                     &                                                                                  \\ \hline
	\end{tabular}}

	\caption{\label{table:table2}%
		The Validation and Test results of \emph{Tool localization} and \emph{hand-tool interaction} results (7-fold Cross-Validation). All values are in therms of $AP_{50}$
	}
	
\end{table*}

\begin{table}
	\centering
	\scalebox{.59}{
		
		\begin{tabular}{|c|l|l|l|l|l|l|l|l|}
			\hline
			& \multicolumn{4}{c|}{\textbf{Not smoothed}}                                                                                                   & \multicolumn{4}{c|}{\textbf{Smoothed}}                                                                                                         \\ \hline
			\textbf{Tool}          & \multicolumn{1}{c|}{\textbf{Precision}} & \multicolumn{1}{c|}{\textbf{Recall}} & \multicolumn{2}{c|}{\textbf{F1}}                            & \multicolumn{1}{c|}{\textbf{Precision}} & \multicolumn{1}{c|}{\textbf{Recall}}   & \multicolumn{2}{c|}{\textbf{F1}}                            \\ \hline
			\textbf{Left hand}     & 0.606                                   & 0.805                                & \multicolumn{2}{l|}{0.692}                                  & 0.693                                   & 0.808                                  & \multicolumn{2}{l|}{0.746}                                  \\ \hline
			\textbf{Right hand}    & 0.942                                   & 0.949                                & \multicolumn{2}{l|}{0.945}                                  & 0.943                                   & 0.959                                  & \multicolumn{2}{l|}{0.950}                                  \\ \hline
			\textbf{Needle driver} & 0.970                                   & 0.946                                & \multicolumn{2}{l|}{0.958}                                  & 0.972                                   & 0.957                                  & \multicolumn{2}{l|}{0.965}                                  \\ \hline
			\textbf{Forceps}       & 0.893                                   & 0.906                                & \multicolumn{2}{l|}{0.900}                                  & 0.909                                   & 0.907                                  & \multicolumn{2}{l|}{0.908}                                  \\ \hline
			\textbf{Scissors}      & 0.842                                   & 0.683                                & \multicolumn{2}{l|}{0.754}                                  & 0.874                                   & 0.755                                  & \multicolumn{2}{l|}{0.810}                                  \\ \hline
			\rowcolor[HTML]{E5E5E5} 
			\textbf{Mean}          & \textbf{0.850}                          & \textbf{0.858}                       & \multicolumn{2}{l|}{\cellcolor[HTML]{E5E5E5}\textbf{0.850}} & \cellcolor[HTML]{E5E5E5}\textbf{0.878}  & \cellcolor[HTML]{E5E5E5}\textbf{0.877} & \multicolumn{2}{l|}{\cellcolor[HTML]{E5E5E5}\textbf{0.876}} \\ \hline
			\rowcolor[HTML]{C0C0C0} 
			\textbf{Accuracy}      & \multicolumn{4}{c|}{\cellcolor[HTML]{C0C0C0}\textbf{0.924}}                                                                                  & \multicolumn{4}{c|}{\cellcolor[HTML]{C0C0C0}\textbf{0.935}}                                                                                    \\ \hline
	\end{tabular}}

	\caption{\label{table:table3}%
		Tool usage task results
	}
	
\end{table}

\section{Motion-based metrics for surgical skills assessment} 

In this section we use our algorithm to develop a fully automatic skills assessment system. In our prior studies we have measured performance using motion metrics such as procedure time, path length, number of hand movements, working volume, etc. \cite{d2015idle,d2016working} In these studies, sensors were used to measure the hand position. Here video data was used to analyze the motion pattern and then derive the metrics. Most sensor systems provide three-dimensional motion data dimensions. Our method relies on two-dimensional projection of the motion data on the image plane. Nevertheless, we get statistically significant separation between medical students and experts.
\subsection{Metric computation}

The three traditional metrics calculated are duration, path length, and number of movements.
The output of the \emph{hand-tool interaction} algorithm is used to define the duration of the procedure. We define the beginning of the procedure as the first frame in which at least one hand is using a tool and the end of the procedure as the last frame in which one of the hands is using a tool. The procedure duration is calculated as the total number of frames divided by 30. The location of an object is defined as the center point of its bounding box. Path length is the two-dimensional distance the hands moved, in the image plane, from the starting point until the end of the suturing task.

The velocity $v$ of each hand is calculated as $v=\sqrt{v_{x}^{2} +v_{y}^{2}}$, where $v_x$ an $v_y$ are the first order numerical derivatives calculated by the centered difference formula on $x$ and $y$ position vectors. The hand is considered as static when the velocity is below the threshold value of $25 pixel / sec $. Finally, the number of movements is defined as the number of times velocity crosses the threshold value divided by two.

In addition to the three traditional metrics in this study, we define two new metrics. Both metrics assess the holding angle of the forceps. The holding angle of the forceps was defined as the aspect ratio of the bounding box of the forceps $width / height$. The first metric is the mean of the aspect ratio throughout the procedure. The second is the standard deviation of the aspect ratio.

\subsection{Results}

\begin{figure}
	\centering
	\includegraphics[scale=0.15]{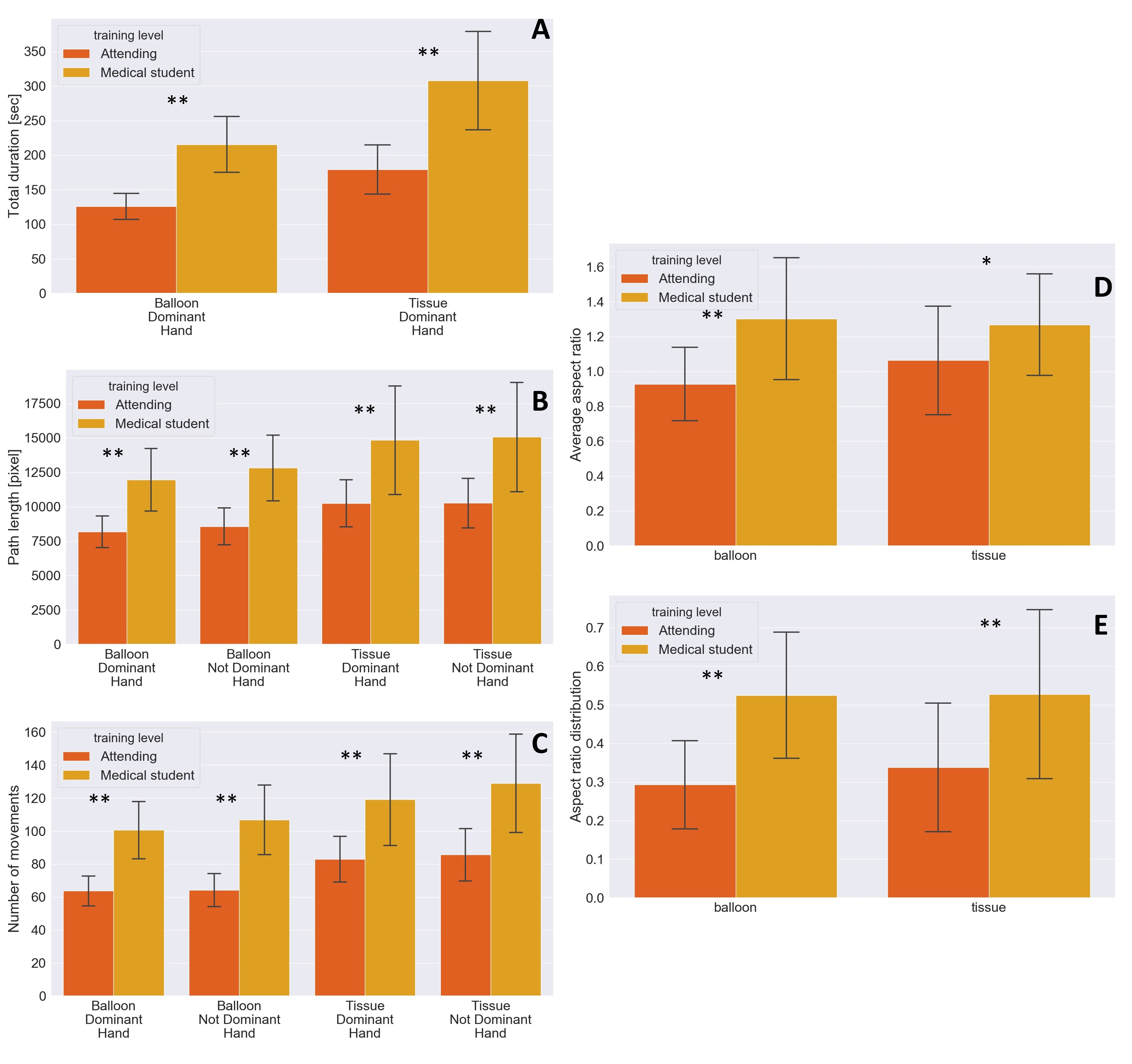}
	\caption{A- procedure duration, B- path length, C- number of movements, D- average mean of the aspect ratio of the forceps' bounding box during its usage, E- average standard deviation of the forceps' bounding box during its usage. $*p-value <0.05$ and $**p-value <0.01$.}
	\label{fig:motion_matrics}
\end{figure}

As anticipated from previous studies, the attending surgeons performed the task in less time, shorter path length, and smaller number of movements [Fig. \ref{fig:motion_matrics}]. The attending surgeons held the forceps at approximately 45$^{\circ}$ while the students held it in a more upright position. Furthermore, lower standard deviation was measured for attending surgeons, suggesting a more stable and consistent grip of the forceps. Since this is a new metric, further analysis is required for providing accurate interpretation of these findings. However, the method might provide new valuable information for the training of new surgeons.

\section{Discussion}

This study focused on the analysis of video data captured using a webcam during open surgery simulation. Our premise was that in open surgery, in addition to detecting the presence of the tools and hands, their interaction needs to be identified. We examined two naive approaches, performing both tasks using one YOLO and using a full network for each task. Since both approaches aimed at solving a standard object detection problem, we would have expected that the results of the integrated system be close to the combined results of the two separate systems. However, the two separate networks yielded significantly better results. A possible explanation for the decreased performance of the integrated network in comparison to the separated networks may be the extended feature overlap between corresponding classes. While using two different networks to solve the two tasks provided good results, this solution is expensive in terms of run time, which is critical for fast simulation analysis. 

Therefore, a dual-task network was constructed and a training scheme was developed. The dual-task network performance was similar to that of two networks, while computational load was only slightly bigger than one network. The system was capable of analyzing approximately 35 frames per second on a NVIDIA Tesla V100 Volta GPU and 15 frames per second on a NVIDIA GeForce GTX 1060 GPU. Thus, feedback may be provided in a timely manner. 

We chose YOLO as our base network due to its short run-time. The network performed very well on our dataset and achieved a mean average precision (mAP) of  87.4 for tool detection. Jin et al. \cite{jin2018tool} trained a slower two-stage object detection network based on Faster R-CNN to detect laparoscopic tools, and reached a mAP of 63.1. The data analyzed in that study was more complex than our data. While our data was captured with a static camera in a simulation environment, in \cite{jin2018tool} the data was the m2cai16-tool-locations dataset, which includes videos taken during cholecystectomy by a mobile endoscopic camera. This suggests that network selection may be based on the combination of run-time limitations and data complexity. This option was explored by Soviany et al. \cite{soviany2018optimizing}. In this study an image difficulty predictor was developed. Based on the assumed difficulty the system decided whether two-stage object detection was required for an image or whether good performance could be achieved with a one stage detector. This approach may balance between run-time limitations and detection requirements.

The dual-task approach developed in this study may be easily expanded to a multi-task approach if needed. A multi-task system could identify multiple structures in an image. For example, in the surgical context, it could help identify the arm-hand interaction and tool-hand interaction as well as identify if the forceps are holding a needle.
This approach is not limited to the surgical arena. In \cite{yoon2019analyzing} basketball movements and pass relationships were studied. For the task, two separate YOLO networks were trained, one for the players (with and without the ball) and one for their jersey number. This could have been done using one dual-task network, saving run-time. 
Another example comes from the field of autonomous driving, where analysis of critical events often depends on object-object interaction between cars, pedestrians, road signs, and other prominent objects, and naturally must be evaluated in real time \cite{herzig2019spatio}.

Video-based motion analysis is a much cheaper and simpler approach than sensor kinematics based assessment. Sensor systems (such as 3D Guidance 6DOF Sensors) may cost thousands of US dollars while in this study a simple webcam was used. In addition, connecting the participants to the sensor system is time consuming. Using RGBD cameras and LIDAR technology may provide a good combination of 3D information, fast setup time, and low prices. In previous works, these technologies were used for action recognition in OR, medical training, and assessment \cite{twinanda2015data,vanvoorst2015fusion}. However, these technologies are not available at most households. Webcams are nowadays a standard component of every computer station or laptop. In addition, due to the COVID-19 outbreak people feel very comfortable in operating this technology. Using a webcam, measuring performance may be as easy as recording a Zoom session. Therefore, the system provides the opportunity for fully automated skill assessment that may be used by the resident or medical student independently. 

Although the system only captured two-dimensional data, our data showed significant differences between experts and novices. Furthermore, in addition to the traditional motion metrics, we identified new metrics that might suggest different techniques for holding the tools and perhaps provide more detailed feedback for improvement. Tool orientation showed that experienced surgeons hold the tool differently than medical students. Moreover, the variability over time of tool orientation was significantly higher for medical students. This suggests they are still exploring the optimal holding technique while experienced surgeons have developed a stable approach. Nevertheless, this is a new metric and more work is required to fully understand it. 

One limitation of our study was that we used the same webcam. For the system to be generalized for domestic use, the DNN should be trained using data from a wide range of cameras. The focus of this current study was to demonstrate that a webcam may be used for assessing technical skill. Thus, data from 13 attending surgeons was captured. This limited our work to a hospital area, and since the focus was comparing experts and novices we kept the system standard. The next phase of our work will be to collect data using multiple systems. This may be done by the medical students using their own equipment.

In this study, we focused on data collected from a medical simulator. While this provides independent merit as an approach for assessing skill and providing automatic feedback, we believe the multi-task approach suggested in this study will be beneficial when analyzing data for more complex simulation or even for the real operating room.

\paragraph{Acknowledgements}Funding for this study was provided by the National Institutes of Health grant 1F32EB017084-01 entitled  "Automated Performance Assessment System: A New Era in Surgical Skills Assessment."

%
\section*{Compliance with ethical standards}

\textbf{Conflict of interest} The authors declare that they have no conflict of
interest.
\\
\\
\textbf{Ethical approval} Study approval was granted by the University of Wisconsin Health Sciences Institutional Review Board and written informed consent was obtained from all participants.
\\
\\
\textbf{Informed consent} Informed consent was obtained from all individual participants included in the study.


\end{document}